\documentclass[preprint,12pt]{elsarticle} 

\usepackage{booktabs}
\usepackage{amssymb}
\usepackage{makecell}
\usepackage{hyperref}
\hypersetup{colorlinks,allcolors=black}
\usepackage{xcolor}
\usepackage{amsthm}
\usepackage{mathtools}
\usepackage{enumitem}
\usepackage{graphicx}
\usepackage{subfigure}
\usepackage{autobreak}
\usepackage{lineno}
\usepackage[margin=2cm]{geometry}
\setcitestyle{authoryear,open={(},close={)}}
\usepackage{multirow}
\usepackage{floatrow}
\usepackage{colortbl}
\usepackage{threeparttable}
\usepackage{empheq}
\usepackage{algorithm}
\usepackage{algorithmic}
\usepackage{float}
\usepackage{times}
\usepackage{amsfonts}

\newfloatcommand{capbtabbox}{table}[][\FBwidth]

\definecolor{orange}{rgb}{1,0.5,0}
\definecolor{red}{RGB}{198,0,35}
\definecolor{amberseldef}{rgb}{1.0, 0.49, 0.0}
\definecolor{ceruleanblue}{rgb}{0.16, 0.32, 0.75}
\definecolor{amber}{rgb}{1.0, 0.49, 0.0}
\definecolor{dodgerblue}{rgb}{0.12, 0.56, 1.0}
\definecolor{pureblue}{rgb}{0, 0, 1.0}

\definecolor{blue}{rgb}{0.0, 0.28, 0.67}

\def\hmath$#1${\texorpdfstring{{\rmfamily\textit{#1}}}{#1}}

\makeatletter
\def\ps@pprintTitle{%
   \let\@oddhead\@empty
   \let\@evenhead\@empty
   \let\@oddfoot\@empty
   \let\@evenfoot\@oddfoot
}
\makeatother

\AtBeginDocument{\hypersetup{citecolor= pureblue,linkcolor = pureblue,urlcolor = dodgerblue}}
\begin{document}
\nolinenumbers

\begin{frontmatter}

\address[1]{Department of Civil Engineering, the University of Hong Kong, Hong Kong, China}
\address[2]{Department of Logistics and Maritime Studies, the Hong Kong Polytechnic University, Hong Kong, China}
\address[3]{Department of Urban and Rural Planning, School of Architecture and Design, Southwest Jiaotong University, Chengdu, China}

\cortext[cor1]{Corresponding author: Email address: \href{sfengag@connect.ust.hk}{sfengag@connect.ust.hk}(Siyuan Feng)}

\author[1]{Taijie Chen}
\author[1]{Zijian Shen}
\author[2]{Siyuan Feng\texorpdfstring{\corref{cor1}}{}}
\author[3]{Linchuan Yang}
\author[1]{Jintao Ke}

\title{Dynamic Adjustment of Matching Radii under the Broadcasting Mode: A Novel Multitask Learning Strategy and Temporal Modeling Approach }

\begin{abstract}
As ride-hailing services have experienced significant growth, the majority of research has concentrated on the dispatching mode, where drivers must adhere to the platform's assigned routes. However, the broadcasting mode, in which drivers can freely choose their preferred orders from those broadcast by the platform, has received less attention. One important but challenging task in such a system is the determination of the optimal matching radius, which usually varies across space, time, and real-time supply/demand characteristics. This study develops a Transformer-Encoder-Based (TEB) model that predicts key system performance metrics for a range of matching radii, which enables the ride-hailing platform to select an optimal matching radius that maximizes overall system performance according to real-time supply and demand information. To simultaneously maximize multiple system performance metrics for matching radius determination, we devise a novel multi-task learning algorithm that enhances convergence speed of each task (corresponding to the optimization of one metric) and delivers more accurate overall predictions. We evaluate our methods in a simulation environment specifically designed for broadcasting-mode-based ride-hailing service. Our findings reveal that dynamically adjusting matching radii based on our proposed predict-then-optimize approach significantly improves system performance, e.g., increasing platform revenue by 7.55\% and enhancing order fulfillment rate by 13\% compared to benchmark algorithms.

\end{abstract}


\end{frontmatter}

\newpage


\section{Introduction}

\begin{figure}[htbp]
  \centering
  \begin{minipage}{\textwidth}
    \centering
    \includegraphics[trim=0cm 10cm 0cm 5cm,width=0.9\textwidth]{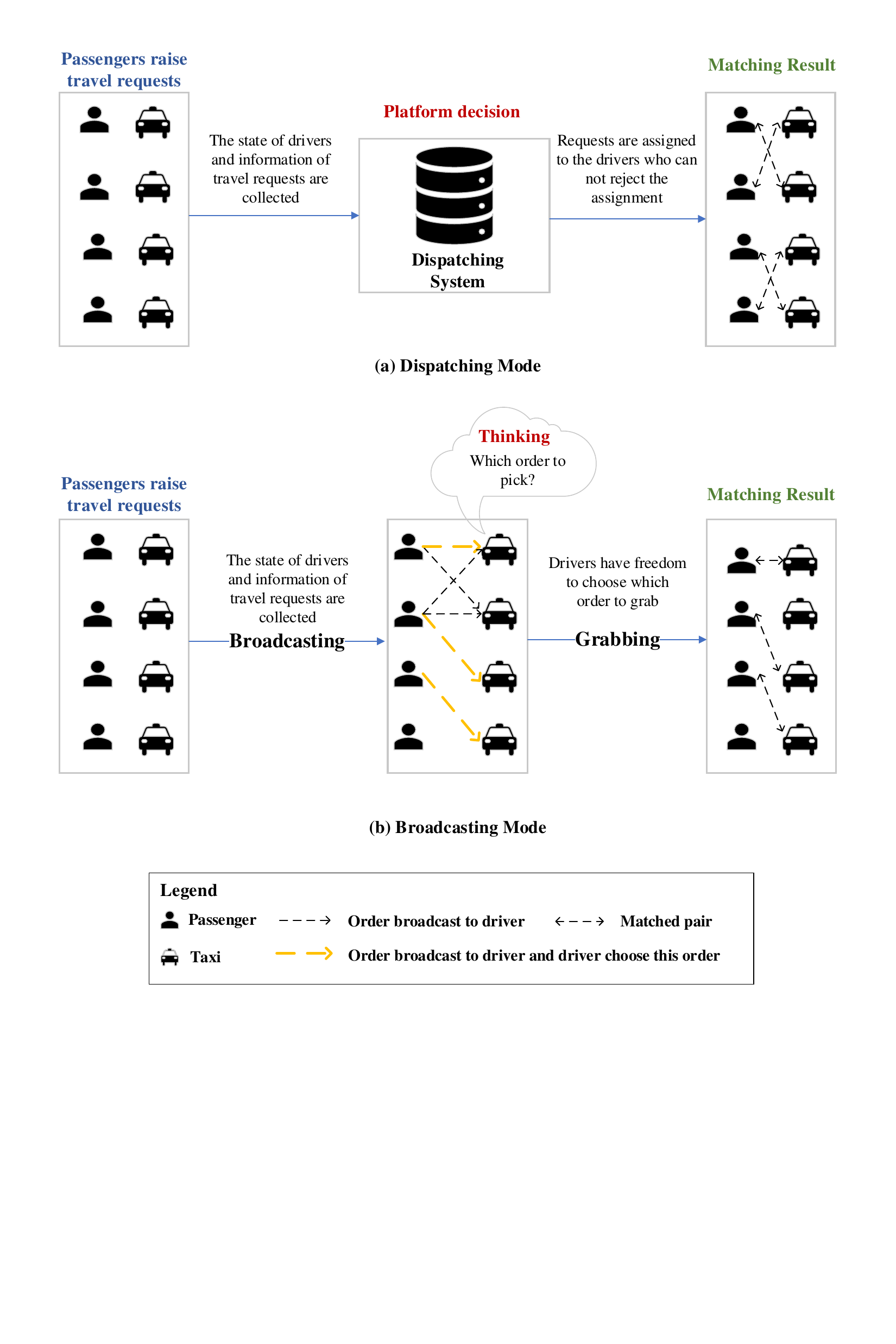}
    \caption{The upper diagram demonstrates the dispatching mode, where a platform assigns a specific order to a driver who must accept it. The lower diagram depicts the broadcasting mode, where orders are sent out to all drivers, granting them the freedom to select their preferred orders.}
    \label{fig:Order matching}
  \end{minipage}
\end{figure}

The matching process between waiting passengers and idle drivers constitutes a critical component in the ride-hailing market, which encompasses two modes: the dispatching mode and the broadcasting mode \citep{yang2020optimizing}. The key distinction between these two modes lies in the degree of freedom accorded to drivers in selecting orders. In the dispatching mode, Transportation Network Companies (TNCs) assign orders to drivers based on specific rules once travel requests are initiated and drivers are not allowed to select which orders to accept \citep{feng2022approximating,ke2020learning}. If drivers choose not to comply with the platform’s assigned orders, they risk facing punitive measures such as being suspended from receiving orders for an extended period. This mode is commonly employed by ride-hailing companies like Didi Chuxing and Uber \citep{wang2019ridesourcing}. As for the second mode, some e-hailing taxi companies, including HK Taxi \footnote{https://hktaxiapp.com/}  and eTaxi \footnote{https://etaxi.com.hk/} in Hong Kong, operate in the broadcasting mode (Figure \ref{fig:Order matching}(b)). In such mode, orders are broadcast to a group of drivers who can then choose whether they want to accept the order based on the information provided by the platform \citep{ashkrof2022ride}, such as estimated fare/distance, origin, and destination of the trip. 

The broadcasting mode offers several noteworthy advantages in the ride-hailing context: 1) It provides drivers with the autonomy to choose which orders they wish to fulfill, thereby safeguarding their rights and interests. 2) It addresses the issue of drivers intentionally choosing indirect routes. A common problem in the current ride-hailing market stems from disgruntled drivers who, reluctant to execute orders assigned by the platform, deliberately prolong passenger wait times by taking detours. This leads to passengers canceling their orders due to excessive waiting time, allowing drivers to evade platform-imposed penalties. Such behavior compromises not only the interests of both passengers and drivers but also the overall efficiency of the system. The introduction of the broadcasting mode can effectively curtail this practice. 3) It has been adopted in several traditional taxi companies' ride-hailing applications and has been integrated into well-established ride-hailing firms like Didi Chuxing's operations. Ride-sourcing platforms can thus switch between the two kinds of modes based on the current needs of markets, improving flexibility and potential operational efficiency. For example, a driver looking to go offline soon might favor shorter orders, while another driver intending to return home might prefer orders leading to destinations closer to their residence.

\begin{figure}[!t]
  \begin{minipage}{0.9\textwidth}
    \centering
    \includegraphics[trim=4cm 5cm 3cm 5cm,width=0.9\textwidth]{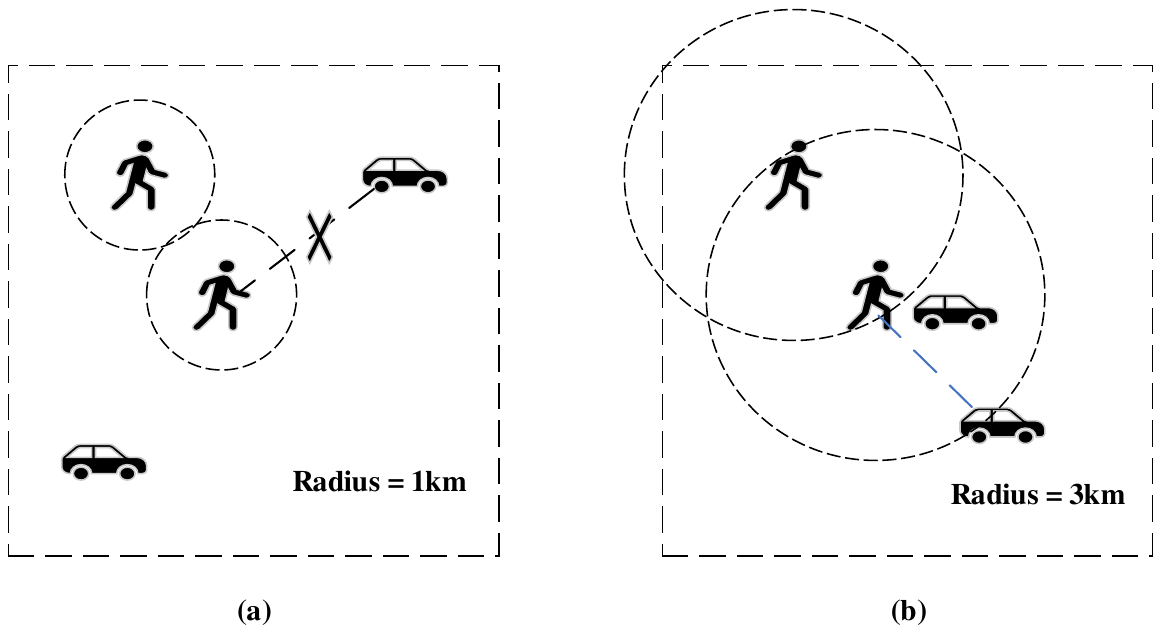}
    \caption{Contrasting Extremes of Broadcasting Radii: Miniscule vs. Vast Scenarios. The left illustration indicates that an excessively small matching radius results in the driver receiving no travel requests. Conversely, the right representation reveals that an overly large matching radius leads to an increased distance between the driver and the passenger.}
    \label{fig:radius extream}
  \end{minipage}
\end{figure}

Although the broadcasting mode is a  very important component for the order matching module in ride-sourcing for taxi services, there is only limited exploration on it \citep{ausseil2022supplier,meskar2023spatio}, especially for the real-time operations. A key operational decision variable is the matching radius, which is a range within which orders can be broadcast to the potential drivers, as shown in Figure \ref{fig:radius extream}. How to determine the appropriate radii is a crucial issue for the broadcasting mode. An excessively small matching radius (Figure \ref{fig:radius extream}(a)) might make it difficult for drivers to receive order requests, while an overly large radius (Figure \ref{fig:radius extream}(b)) may require drivers to spend substantial time picking up passengers. The varying demand and supply distribution across space further increases the complexity of the problem, meaning different corresponding matching radii should be set for different locations simultaneously. Moreover, the on-demand mobility market is highly dynamic, making the radii to be also decided dynamically.

To address the research gap, we introduce the Dynamic Broadcasting Radii Adjustment System (DBRAS), integrating historical data with real-time supply-demand dynamics to adjust broadcasting radii (also called matching radii). Specifically, our Dynamic Broadcasting Radii Adjustment System (DBRAS) incorporates inputs such as 1) real-time market states, including the number of idle vehicles and travel requests, 2) historical market states and performance metrics, such as order fulfillment rate, driver utilization rate, average pickup distance, and platform revenue, and 3) various broadcasting radii. The system then predicts performance metrics including order fulfillment rate, driver utilization rate, average pickup distance, and platform revenue. Based on these predictive results, we can select the broadcasting radius that yields the most optimal performance metrics. The strength of our system lies in its ability to decipher hidden spatial-temporal patterns and trends, a feat unattainable by traditional optimization algorithms. To be more specific, our system is composed of two parts: 1) We have devised a temporal prediction model that employs the encoder of the Transformer \citep{vaswani2017attention} to encode supply, demand, and other market factors. The encoded results are then fed into a fully connected layer to yield predicted performance metrics. 2) We conceptualize the prediction of multiple performance metrics as a multi-task prediction problem, based on which we pick the most appropriate matching radii. Through extensive simulation-based experiments in Hong Kong and Manhattan, we demonstrate the effectiveness and efficiency of the proposed predict-then-optimize approach. The major contributions of this study are summarized as follows:

\begin{enumerate}[label=(\alph*)]
    \item We make the first attempt to develop a predict-then-optimize framework integrated with deep learning approaches to dynamically adjust matching radius for system performance optimization in ride-hailing systems with the broadcasting mode. 
    \item We develop a novel multi-task training strategy to balance the tradeoff among the optimization for various system performance metrics in ride-hailing systems, including order fulfillment rate, driver utilization rate, average pickup distance, and platform revenue. The multi-task learning training strategy is shown to be effective in attaining much faster convergence speed and converging to a lower loss. 
    \item We have conducted extensive experiments based on a tailored simulation platform for the broadcasting mode operations, which validate the effectiveness of our proposed predict-then-optimize approach and multi-task training strategies.
\end{enumerate}

The structure of the remainder of this paper is as follows. Section 2 offers a concise literature review, shedding light on the order matching process, temporal modeling, and multi-task learning within the context of the ride-sourcing market. In Section 3, we define the problem of the dynamic adjustment of the broadcasting radius, elaborating on the variables at play within this problem. Following this, Section 4 first presents a comprehensive introduction to our proposed system for dynamically adjusting the radius, before diving into two pivotal components of the system: 1) the performance metric prediction model based on Transformer, and 2) the multi-task training strategy. In Section 5, we delineate the experimental settings, datasets, benchmark models, and evaluation metrics. Moreover, we showcase the training results of the prediction model, as well as the outcomes of testing the dynamic radius adjustment system within our simulator. A detailed analysis of these experimental results is also provided. Lastly, Section 7 encapsulates the contents of the paper, while also casting a gaze towards future research avenues in the field.

\section{Literature Review}
\subsection{Order Matching}
In the current era marked by rapid advancements in mobile internet and wireless communication technologies, the ride-sourcing market has experienced significant growth. At the core of ride-hailing platform operations lies the essential process of order matching. Passengers, using intelligent mobile devices and a range of apps, request transportation, which initiates a waiting period for vehicle pairing. This order-matching process typically follows one of two modes: the dispatching mode and the broadcasting mode. In the dispatching mode, the ride-sourcing platform assigns order requests to idle drivers, and drivers cannot reject the assignment if they do not want to be punished by the platform. A significant body of academic research has explored this mode. The studies by \cite{liu2022machine} and \cite{qin2022reinforcement} present a thorough review of these methodologies. The dispatching process can take into consideration a variety of objectives. For instance, \cite{xu2018large} focus on maximizing drivers' revenue, \cite{ozkan2020dynamic} aim to maximize the number of matched pairs, while \cite{wong2006optimal} strive to minimize the passengers' average waiting time. Early research typically employed static and greedy matching strategies, such as pairing the nearest drivers and passengers \citep{bailey1987simulation}. Currently, the utilization of reinforcement learning to solve sequential decision-making problems is gaining traction. The model introduced by \cite{haliem2021distributed} optimizes the matching, pricing, and vehicle dispatching collectively. \cite{ke2020learning} proposed a model that integrates reinforcement learning with integer programming to delay passenger matching, thereby reducing the average pickup time. Some researchers have leveraged reinforcement learning techniques to optimize the market by integrating the dispatching process with public transit services \citep{feng2022coordinating}. Conversely, the broadcasting mode broadcasts orders to drivers, who are then free to select the orders they wish to accept. In the broadcasting mode, drivers face no penalties for rejecting an assignment, a stark contrast to the dispatching mode. \cite{ashkrof2022ride} posit that pickup time (the travel time from the driver's location to the passenger's origin) negatively impacts ride acceptance. \cite{ausseil2022supplier} depict the matching process for peer-to-peer transportation platforms as a two-stage decision problem. The first decision relates to the provision of supplier menus, and the second decision involves demand assignment post-supplier selection. Concurrently, they put forth a multiple-scenario approach to address the aforesaid two-stage decision problems. Finally, \cite{sun2020taxi} implemented an integer programming method to allocate customer requests between two systems (Inform system and Assign system) and to determine the maximum matching radii.

Despite its role as an operational variant within the ride-hailing market, the broadcasting mode and its spatial-temporal patterns remain relatively underexplored. This study aims to address this research gap. We plan to employ temporal modeling to optimize the ride-hailing market operating under the broadcasting mode by manipulating the matching radius.

\subsection{Temporal Modeling}
Temporal modeling refers to the methods that consider the importance of the time order of observations, which is widely used in the fields of machine learning, finance, and economics, among others. It has also been widely applied in the field of intelligent transportation. \cite{ahmed1979analysis} was the first to employ the autoregressive integrated moving average (ARIMA) model for predicting traffic flow, which subsequently led other researchers to explore variations of the ARIMA model for related studies \citep{levin1980forecasting, hamed1995short, billings2006application}. In recent years, with the rapid advancement of deep learning, Recurrent Neural Networks (RNN) and Long Short-Term Memory (LSTM) models have been increasingly integrated into transportation research, encompassing demand and supply forecasting \citep{xu2017real, ke2017short}, and traffic flow prediction \citep{wu2016short, yu2017spatiotemporal}, among others. The attention mechanism, having demonstrated superior performance in time series forecasting, has also attracted considerable scholarly interest, with many researchers adopting Transformer-based models for time series forecasting. \cite{ramana2023vision} used a Transformer to predict urban area traffic congestion. \cite{xie2022multisize} proposed a novel MultiSize Patched Transformer approach, which incorporates rich and unified context, to predict crowd flows. \cite{xu2023multi} used the encoder of the Transformer to predict demand and supply in bike-sharing systems. 

However, previous research typically used temporal models to predict some traffic states like traffic flow, supply, demand, etc. Unlike these studies, we aim to forecast aggregate performance indicators over a future period after adopting different radii simultaneously. These indicators include order fulfillment rate, average pickup distance, platform revenue, driver utilization rate, etc. Based on the forecasted performance indicator values, we can select the radii value to be adopted directly rather than conducting other optimization procedures.

\subsection{Multi-task Learning}
Transportation systems exhibit a high degree of complexity, with a multitude of traffic states that reciprocally influence each other. Predicting a single state often falls short of addressing the intricate challenges posed by these complex transportation networks. Therefore, there is a pressing need for the simultaneous prediction or optimization of multiple traffic states. The field of machine learning has responded to this need by proposing multi-task learning methods, which are designed to handle the complexities of predicting multiple targets. This approach has also resonated with researchers in the transportation field, with many adopting multi-task learning methodologies in their studies. For instance, \cite{ke2021joint} utilized multi-graph and multi-task learning to predict demand across varied service modes. \cite{9376697} utilize factorized Graph Neural Networks to co-predict the Zone-Based and OD-based demand. Similarly, Zhang et al. \cite{zhang2016dnn} introduced an architecture that ingeniously combines spatial-temporal and global components to predict zone-level travel demand. Likewise, \cite{wang2022multitask} proposed a Multi-task Hypergraph Convolutional Neural Network, a novel approach to handle the heterogeneity introduced by disparate source data. \cite{liang2023region} utilize multi-task adaptive graph attention network to predict the region-level demand.

However, previous multi-task research in the field of transportation mostly used fixed weights for different tasks and do not design a task-specific multi-task learning strategy. In this study, we propose several multi-task learning strategies that allow for the dynamic adjustment of weights for different tasks during the training process.

\section{Preliminaries}
In this section, we will define the performance metrics prediction problem, and weights update for each task in the multi-task problem, while introducing the notation used for variables in our study.

\textbf{Definition 1} \textit{(Dividing Time and Regions)}: Our study area is a square region, divided into an \(\mathit{I*I}\) grid, resulting in \(\mathit{I^2}\) zones. We use indices ranging from \(\mathit{0}\) to \(\mathit{I^2-1}\) to denote the grids. For instance, the \(\mathit{i}\)-th grid is denoted as \(\mathit{g_i}\). We employ statistical performance metrics within a five-minute time window in this study, which include the order matching rate, the number of idle drivers, the number of unmatched orders, and the average pickup distance between matched orders and drivers, among other data.

Building on Definition 1, we provide definitions for the following variable:

(1) Time-of-day

Our research partitions the 24-hour day into four-time segments: morning, evening, midnight, and other periods, with morning and evening classified as peak hours. We use the dummy variable ($h_t$) to denote the time-of-day variable.

\begin{equation}
h_t =
\begin{cases}
    0 & \text{if } t \text{ belongs to evening,} \\
    1 & \text{if } t \text{ belongs to morning,} \\
    2 & \text{if } t \text{ belongs to midnight,} \\
    3 & \text{if } t \text{ belongs to other periods.}\\
\end{cases}
\end{equation}

\begin{figure}[!t]
  \centering
  \begin{minipage}{\textwidth}
    \centering
    \includegraphics[trim=4cm 2cm 7cm 0cm,width=0.6\textwidth]{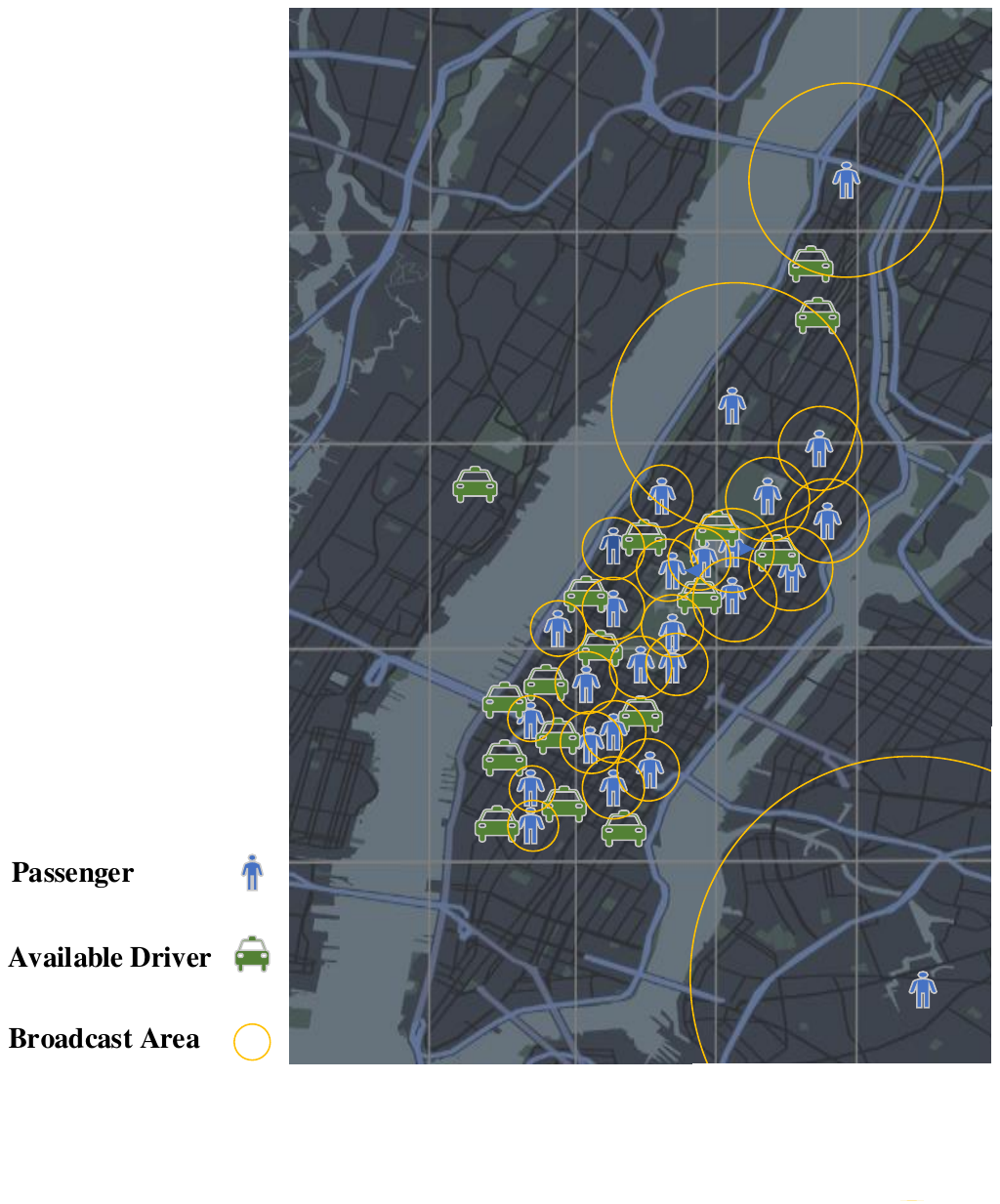}
    \caption{Broadcasting Radii in Different Region}
    \label{fig:broadcast_radius}
  \end{minipage}
\end{figure}

\textbf{Definition 2} \textit{(Broadcasting Mode)}: Figure \ref{fig:broadcast_radius} illustrates the concept of the broadcasting mode, where the blue humanoid icons represent passengers with travel requests, and the green car icons represent available taxis. The yellow region centered around the passenger represents the broadcasting area. The broadcasting radius, denoted by \(\mathit{R}\), corresponds to a circular region centered at the order's location, with a radius of \(\mathit{R}\) from the order's position. The broadcasting radius for grid \(\mathit{g_i}\) at time \(\mathit{t}\) is represented as \(R_t^i\), where \(R_t^i \in \mathbb{R^+}\). Drivers within the broadcasting area can compete for the order during bidding rounds. 

The performance metrics like order fulfillment rate, average pickup distance, driver utilization rate, and platform revenue are denoted as $o$, $d$, $u$, $p$, respectively, where the nearest market information like number of idle vehicles (supply), number of order requests (demand), number of total vehicles (including occupied and unoccupied vehicles) denoted as $N_e$, $N_o$, $N_v$ respectively.

\textbf{Problem 1}
For each grid $g_i$, given the previous observations and pre-known information

\(\{{N_e}_{s}^i,{N_o}_s^i,{N_t}_s^i|s=0,\dots,t;{o}_s^i,{d}_s^i,{u}_s^i,{p}_s^i|s=0,\dots,t-1;R_s^i|s=0,\dots,t-1|;g_i;h_t\}\) combining k-th value $\hat{R_t^{ik}} \in \mathbb{R^+}$, denoted as $\mathbf{x} \in \mathbb{R}^{T \times D}$,  
predict \(\{\hat{{o}_t^i},\hat{{d}_t^i},\hat{{u}_t^i},\hat{{p}_t^i\}}\) denoted as $\hat{y}_{t}^{ik}$, then according the best $\hat{y}_{t}^{i*}$ to determine the optimal radius as $R_t^{i*}$. 

\textbf{Definition 3} \textit{(Multi-task Learning)}: In our research, we aim to simultaneously predict the values of $o$, $d$, $u$, and $p$. We treat this problem as a multi-task problem composed of four tasks, each sharing a common backbone and possessing a unique individual layer. A weight $w$ is assigned to each task, signifying its impact on the shared backbone. The weight for the i-th task at the t-th training step is denoted as $w_t^i$, such that $\sum_{i=1}^{4}w_t^i = 1$. This ensures that the cumulative weights of all tasks at any given training step equal 1, maintaining a balanced influence from each task on the shared backbone. The loss at the t-th training step is indicated as $l_t$, and for the i-th task, the loss is denoted as $l_t^i$. We multiply the weights by their corresponding losses to calculate the gradient. Subsequently, we use gradient back-propagation to update the neural network's parameters. This methodology enables our model to learn and adapt across multiple tasks, optimizing prediction performance for $o$, $d$, $u$, and $p$. A crucial aspect of this process involves determining the weight of each task at every training step. This challenge prompts us to propose Problem 2.

\textbf{Problem 2}
For the t-th training step, given the previous T steps losses $[l_{t-T},\dots,l_{t}]$, then determine the $w_{t+1}^i$ for each task.

\section{Methodology}
In this study, we aim to determine the suitable radius for each grid during each time interval throughout the examined period. Our objective is to simultaneously maximize platform revenue, order fulfillment rate, and driver utilization rate and minimize the average pickup distance between the drivers' location and the passengers' boarding position. To achieve real-time determination of the best matching radius for each area, we employ a temporal neural network. This allows us to understand the influence of historical states on the current market, predict the performance metric values under given conditions (such as the number of idle vehicles and the volume of request orders), and thereby determine the optimal radius by choosing the radius which can maximize the performance metrics. Given the multi-faceted nature of our optimization goal, it's necessary to predict the values of multiple performance metrics concurrently. As such, our prediction task is expanded to encompass multiple objectives. To better balance the weight of each task within this multi-objective prediction, we propose several multi-task learning strategies. These strategies factor in historical losses, ensuring improved performance across all prediction tasks. To facilitate improved simulation testing, we employ logistic regression to model historical data, thereby representing the driver's behavior in broadcast mode. Our Dynamic Broadcasting Radii Adjustment System (DBRAS) is comprised of all the modules mentioned above, working together seamlessly. The structure of this paper is as follows: Section 4.1 provides an overview of the dynamic radius adjustment system and discusses the interplay between each module. Section 4.2 explains the use of logistic regression to model drivers' order-grabbing behaviors. Section 4.3 introduces the temporal prediction model, and Section 4.4 details our multi-task learning strategy.

\begin{figure}[!b]
  \begin{minipage}{1\textwidth}
    \centering
    \includegraphics[trim=0cm 2cm 1cm 1cm,width=0.9\textwidth,]{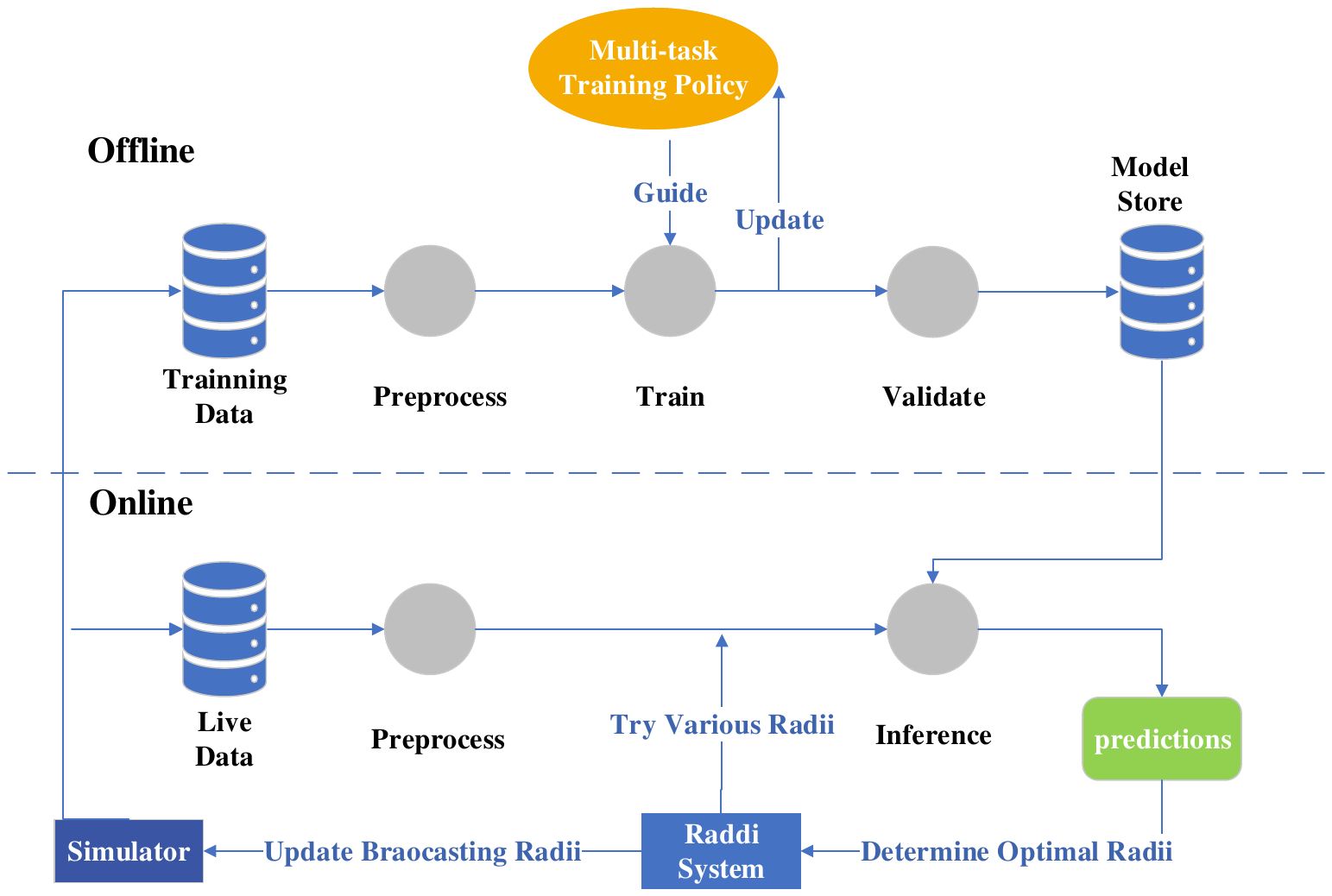}
    \caption{Overview of DBRAS}
    \label{fig:dbrs_structure}
  \end{minipage}
\end{figure}
\subsection{Dynamic Broadcasting Radii Adjustment System}
In this section, we provide an overview of the Dynamic Broadcasting Radii Adjustment System (DBRAS), depicted in Figure \ref{fig:dbrs_structure}. Understanding the interplay between its submodules will facilitate a comprehensive grasp of the system. The DBRAS comprises two key components: offline learning and online radius adjustment. The specifics for each part are as follows:
\begin{itemize}
    \item \textbf{Offline Learning:} In the offline process, we first collect training data from the simulator. This data, which includes features and labels, is then normalized in the preprocessing process prior to the training process. The processed data is subsequently fed into our proposed Transformer-encoder-based (TEB) model for the e-hailing broadcasting system, which is based on a Transformer-encoder. Notably, to enhance multi-task training performance, we introduce a multi-task training strategy named the Weighted Exponential Smoothing Multi-task (WESM) learning strategy. Finally, we store the optimal model for online use.
    \item \textbf{Online Radius Adjustment:} In the online process, we first develop a simulator incorporating the driver-grabbing-order behavior model. This serves to accurately represent the ride-sourcing market under the broadcasting mode. Subsequently, operation data collected from the simulator is normalized in the preprocessing procedure. This processed data, in combination with different radii, constitutes the input for the stored prediction model. Through this model, performance metrics for different radii are generated. Ultimately, the Radii System selects the radius that results in the best-predicted performance metrics, e.g., in our experiments, we select the radius with the maximum value of the summarization of all performance metrics. This optimal broadcast radius is then set as the matching radius for the subsequent round of simulation.
\end{itemize}

Through these two components, the DBRAS shows its robust capacity to learn from historical data and adjust broadcasting radius in real-time, making it a valuable tool for optimizing ride-hailing services.

\subsection{Driver-grabbing-order Behavior Model}
Simulating a ride-sourcing market under the broadcasting mode necessitates a realistic depiction of drivers' order-grabbing behaviors. This section elucidates how we model such behaviors in our simulation. Intuitively, the propensity of drivers to accept an order is likely positively correlated with the order's value and negatively correlated with the distance between the driver and the passenger. With this understanding, we employ a logistic regression model, trained on driver trajectory data provided by eTaxi, to simulate driver behavior. The model's first input, represented by $x_1$, is the pickup distance. The second input, denoted by $x_2$, is the price of the order. The term $\epsilon$ symbolizes the noise factor, which we assume follows a standard normal distribution. This model allows us to predict the likelihood of drivers accepting an order, given the order's value and pickup distance, thereby enabling a more accurate simulation of the ride-sourcing market under the broadcasting mode. Therefore, the probability of accepting the order is given by
\begin{equation}
P(Y=1 | X) = \frac{1}{1 + e^{-(\beta_0 + \beta_1 x_1 + \beta_2 x_2 +  \epsilon) } }
\label{eq:logistic_regression}
\end{equation}
The loss function is given by
\begin{equation}
\mathcal{L}(y, \hat{y}) = -[y \log(\hat{y}) + (1 - y) \log(1 - \hat{y})]
\end{equation}
where \(\boldsymbol{\mathit{y}}\) is the actual value and \(\boldsymbol{\mathit{\hat{y}}}\) is the predicted value.
\subsection{Transformer-Encoder-Based model for the E-hailing Broadcasting system}
In this section, we introduce the Transformer-Encoder-Based (TEB) model for an e-hailing broadcasting system. The Transformer, as introduced by \citep{vaswani2017attention}, is a renowned encoder-decoder structure built upon a self-attention mechanism. Its efficiency outperforms that of LSTM and RNN due to its capacity for parallel processing. Moreover, the transformer's unique ability to handle long-distance dependencies allows it to establish links between time steps, irrespective of the distance between them. These distinctive features let us incorporate the Transformer encoder as the backbone of the TEB. A comprehensive scheme of the TEB can be viewed in Figure \ref{fig:model structure}. Inputs into the TEB are formed by the concatenation of demand, supply, and other influential factors, represented as $\mathbf{x}$. The encoder itself is composed of several discrete blocks. Each block within the Transformer encoder is constituted by a self-attention layer, an MLP layer, and a LayerNorm layer. 'SelfAtten' signifies the projection from the input to the output within the self-attention layer. 'MLP' denotes the transformation from the input to the output within the MLP layer. Lastly, 'LayerNorm' symbolizes the projection from the input to the output within the layer normalization. The mathematical formulation of these functions is shown in Appendix. The transformation process within the Transformer encoder, converting input to output, is defined as follows:
\begin{figure}[!b]
  \centering
  \begin{minipage}{\textwidth}
    \centering
    \includegraphics[trim=0cm 0cm 0cm 0cm, width=1\textwidth]{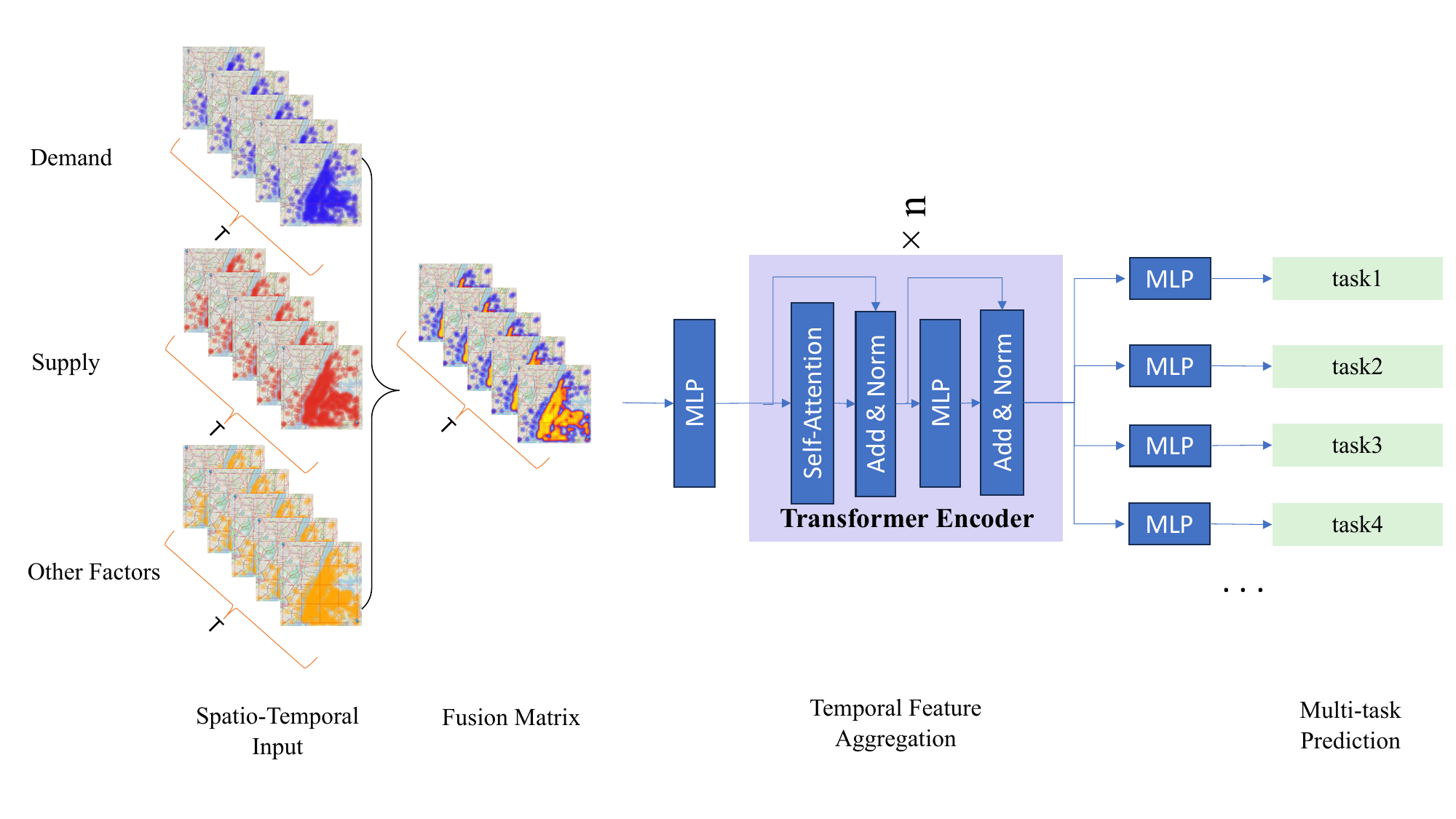}
    \caption{The structure of TEB model}
    \label{fig:model structure}
  \end{minipage}
\end{figure}

\begin{equation}
    \mathbf{o^{'}} = MLP(\mathbf{x})
    \label{eq:f_mlp}
\end{equation}
\begin{equation}
   \mathbf{h_{1}^{'}} = SelfAtten(\mathbf{o^{'}} + LayerNorm(\mathbf{o^{'}}))
    \label{eq:temporal model1}
\end{equation}
\begin{equation}
    \mathbf{h_{2}^{'}} = MLP(\mathbf{h_{1}^{'}} + LayerNorm(\mathbf{h_{1}^{'}}))
    \label{eq:temporal model2}
\end{equation}
\begin{equation}
    \mathbf{o_{t}^{'}} = \mathbf{h_{2}^{'}}
    \label{eq:temporal model3}
\end{equation}
where $o^{'}$ denotes the input fed into the Transformer encoder, while $h^{'}$ and $h^{'}$ represent the hidden states within the encoder. Equations \ref{eq:temporal model1} ,\ref{eq:temporal model2} and \ref{eq:temporal model3} are iteratively processed $n$ times. After the transformations within the Transformer encoder, the output is relayed to several MLP layers, culminating in the final predictions for a range of tasks.
\begin{equation}
    \hat{y}^{i} = MLP^{i}(\mathbf{h_{2}^{'}}), i \in N = \{x \in \mathbb{Z}:1 \leq x \leq m\} 
    \label{eq:fc model}
\end{equation}
The goal during the training phase is to optimize the parameters of the TEB model, symbolized as $\theta$, to minimize the Mean Squared Error (MSE) loss. This objective can be represented as:

\begin{equation}
\min_{\theta} \: \mathcal{L}(\theta) = \frac{1}{c} \frac{1}{m}  \sum_{j=1}^{c} \sum_{i=1}^{m}  (y_{j}^{i} - \hat{y}_{j}^{i})^2
\end{equation}
In this context, $y^{i}$ stands for the actual value of the $i$-th task for the $j$-th instance, $\hat{y}^{i}$ designates the predicted value of the $i$-th task for the same instance, $m$ signifies the total number of tasks, and $c$ corresponds to the count of instances.

\subsection{Multi-task Learning}
In this section, we present our devised multi-task training strategy, termed the Weighted Exponential Smoothing Multi-task (WESM) training strategy. WESM is engineered to expedite the convergence of our proposed TEB model and ensure it settles at a lower loss. To optimize the training of our multi-task prediction model, we initially propose the Adaptive Multi-task (AM) training strategy (as shown in Algorithm \ref{alg:am policy}). AM aims to equalize the learning process across multiple tasks in a time-dependent scenario, a concept further expounded in Section 5. The central premise of this strategy is to prioritize tasks with the highest cumulative loss over time and modify their model parameters accordingly. However, AM assumes that all historical losses contribute equally to the current parameters' update, and it can only modify the parameters of a single prediction task per training iteration. To address these constraints of AM, we ultimately propose WESM. This strategy posits that the influence of historical losses decreases as the distance from the current training step expands. Concurrently, WESM has the capacity to update the parameters of multiple prediction tasks in a single step. 
\begin{algorithm}[H]
\caption{Adaptive multi-task Training Strategy}
\label{alg:am policy}
\begin{algorithmic}[1]
\REQUIRE Features: $\mathbf{x}$;
Labels: $\mathbf{y}$;
time step: $t$;
Historical losses: $\mathbf{L}$.
\ENSURE Updated task with maximum aggregated loss
\FOR{$i = 1$ to $m$}
    \STATE Predict the value of task $i$ which is $\hat{y}_{t}^{i}$
    \STATE Calculate loss for Task $i$: $l_{t}^{i} = (y_{t}^{i} - \hat{y}_{t}^{i})^2$
    \STATE Calculate aggregated loss for task $i$: $$L_{t}^{i} =  \frac{\sum_{j=1}^{T} \gamma^i * \mathbf{L}_{t-j}^{i} +  l_t^i}{T + 1}$$
\ENDFOR
\STATE Determine the task with the maximum aggregated loss: $i* = \mathop{\arg\max}\limits_{i} L_t^{i} $
\STATE Update the task i* with the maximum aggregated loss : $L_{t}^{i*} $
\end{algorithmic}
\end{algorithm}
The intricate steps of the Weighted Exponential Smoothing Multi-task (WESM) training strategy are outlined in Algorithm \ref{alg:wes policy}. The inputs to this algorithm comprise predictors (denoted as $\mathbf{x} \in \mathbb{R}^{T \times D}$), performance metrics (denoted as $\mathbf{y} \in \mathbb{R}^{T \times m}$), a smoothing factor (denoted as $\gamma$), a time step $t$, and historical losses (denoted as $\mathbf{L} \in \mathbb{R}^{T \times m}$) of the preceding $T$ training steps. Next, we delve into the particulars of the algorithm. Initially, we compute the normalization factor $\overline{n}$ as the sum of a geometric series, with $\gamma$ being the common ratio. This factor ensures that the influence of historical losses on the current parameters update is scaled suitably, taking into account both the decay factor and the number of time steps. Subsequently, the aggregated loss $L_t^i$ for the $i$-th task is calculated as $\frac{\sum_{j=1}^{T} \gamma^i * \mathbf{L}_{t-j}^{i} + l_t^i}{\overline{n}}$, where $i \in \mathbb{Z}:0 \leq j \leq T$. Here, $\mathbf{L}_{t-j}^{i}$ denotes the loss for the $i$-th task at the $(t-j)$-th training step, and $l_t^i$ signifies the loss at the current training step. Following this, the decay cumulative loss of each task is divided by the total decay cumulative losses of all tasks to determine the weight $w_t^i$ of each task's parameter update during the training process. Finally, WESM employs the weight calculated for each task to adjust the corresponding parameters in the multi-task prediction model.

\begin{algorithm}[H]
\caption{Weighted Exponential Smoothing Multi-task Training Strategy}
\label{alg:wes policy}
\begin{algorithmic}[1]
\REQUIRE Features: $\mathbf{x}$;
Labels: $\mathbf{y}$;
Smoothing factor: $\gamma$;
time step: $t$;
Historical losses: $\mathbf{L}$.
\ENSURE Update all tasks simultaneously with calculated weights.
    \STATE Compute the normalization factor: $\overline{n} = \sum_{i=0}^{T} \gamma^i  = \frac{1-\gamma^{T+1}}{1-\gamma}$.
    \FOR{$i = 1$ to $m$}
    \STATE Predict the value of task $i$ which is $\hat{y}_{t}^{i}$
    \STATE Calculate the loss at step $t$: $l_t^i = MSE(y_t^i,\hat{y}_t^i)$
    \STATE Calculate aggregated loss for task $i$: $$L_{t}^{i} =  \frac{\sum_{j=1}^{T} \gamma^i * \mathbf{L}_{t-j}^{i} +  l_t^i}{\overline{n}}$$

    \ENDFOR
    \FOR{$i = 1$ to $m$}
    \STATE Compute the weight for task $i$: $$
w_{t}^{i} = \frac{L_{t}^{i}}{\sum_{j=1}^{n} L_{t}^{j}} 
$$
    \STATE The $i$-th task uses weight $w_{t}^{i}$ to perform backpropagation to update model parameters. 
    \ENDFOR

\end{algorithmic}
\end{algorithm}

\section{Experiments}
\subsection{Setup and baselines}
\begin{figure}[!t]
\centering
\subfigure[The investigated region around Manhattan]{
\begin{minipage}[t]{0.5\linewidth}
\centering
\includegraphics[trim=0cm 0cm 0cm 0cm,width=2.5in,height=2.5in]{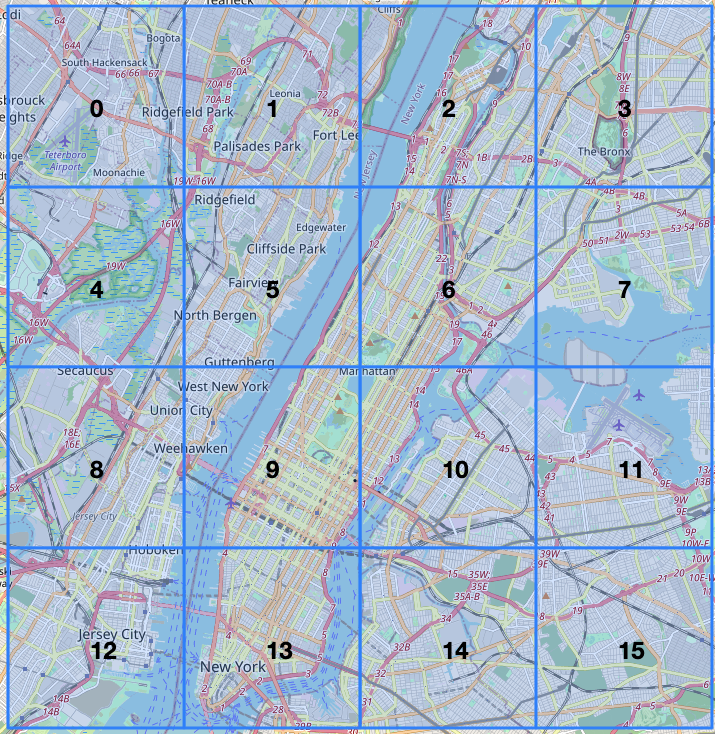}
\end{minipage}%
}%
\subfigure[The investigated region in Hong Kong Island]{
\begin{minipage}[t]{0.5\linewidth}
\centering
\includegraphics[trim=0cm 0cm 0cm 0cm,width=2.5in, height=2.5in]{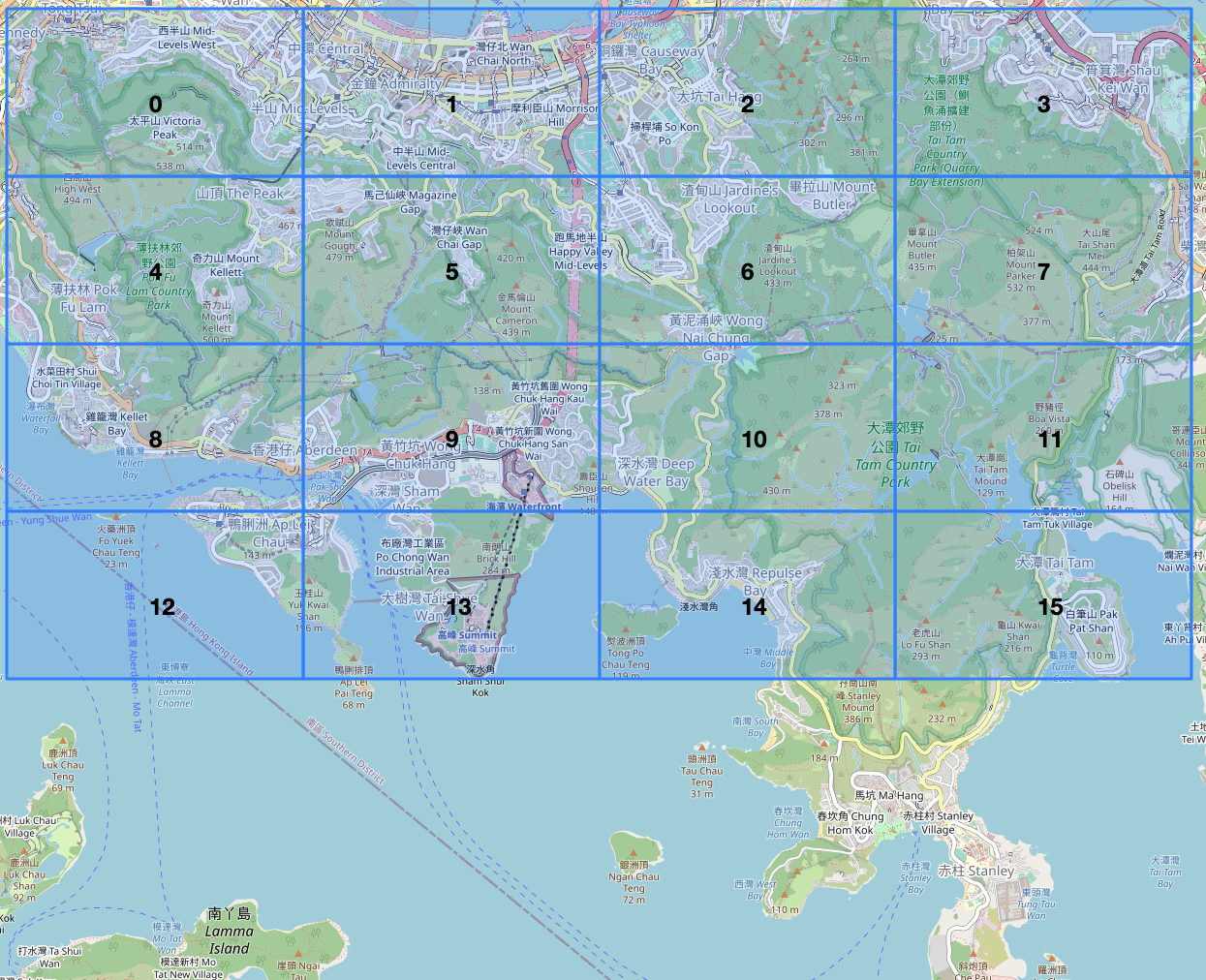}
\end{minipage}%
}%

\centering
\caption{The region under investigation, depicted by the blue shaded area, is partitioned into a 4x4 grid. Each grid cell is assigned a label ranging from 0 to 15.}
\label{fig: research_area}
\end{figure}

\begin{figure}[!t]
\centering
\subfigure[Demand in different hours at Manhattan]{
\begin{minipage}[t]{0.5\linewidth}
\includegraphics[trim=0cm 0cm 0cm 0cm,width=3.3in]{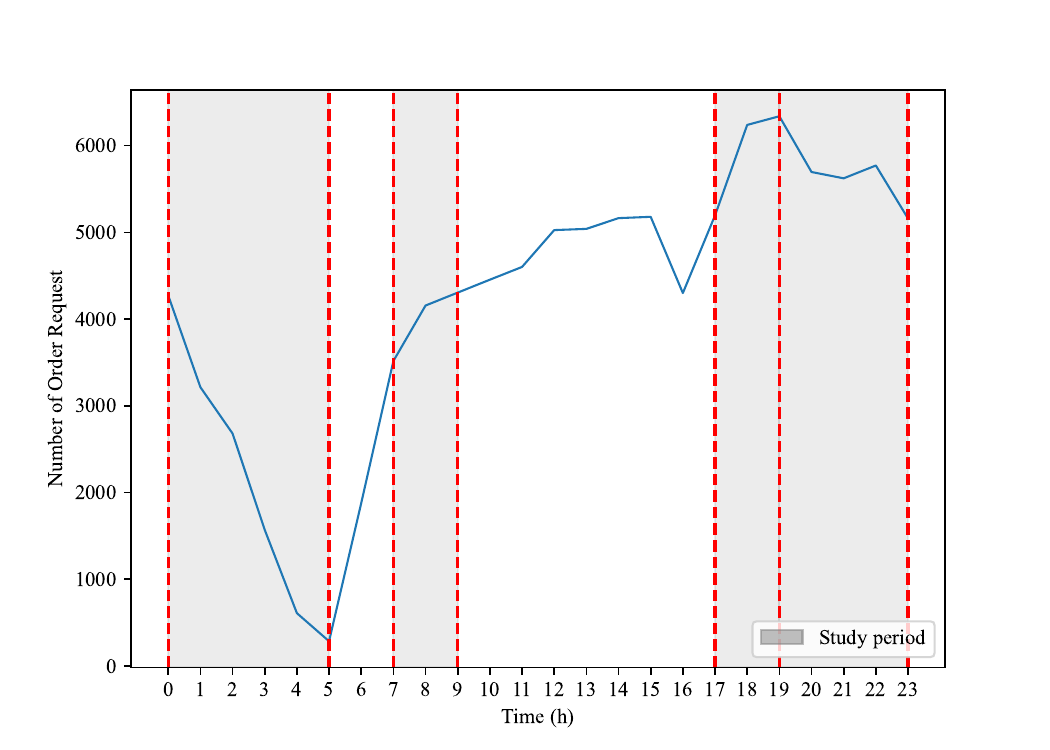}
\end{minipage}%
}%
\subfigure[Demand in different hours at Hong Kong Island]{
\begin{minipage}[t]{0.5\linewidth}
\includegraphics[trim=0cm 0cm 0cm 0cm,width=3.3in]{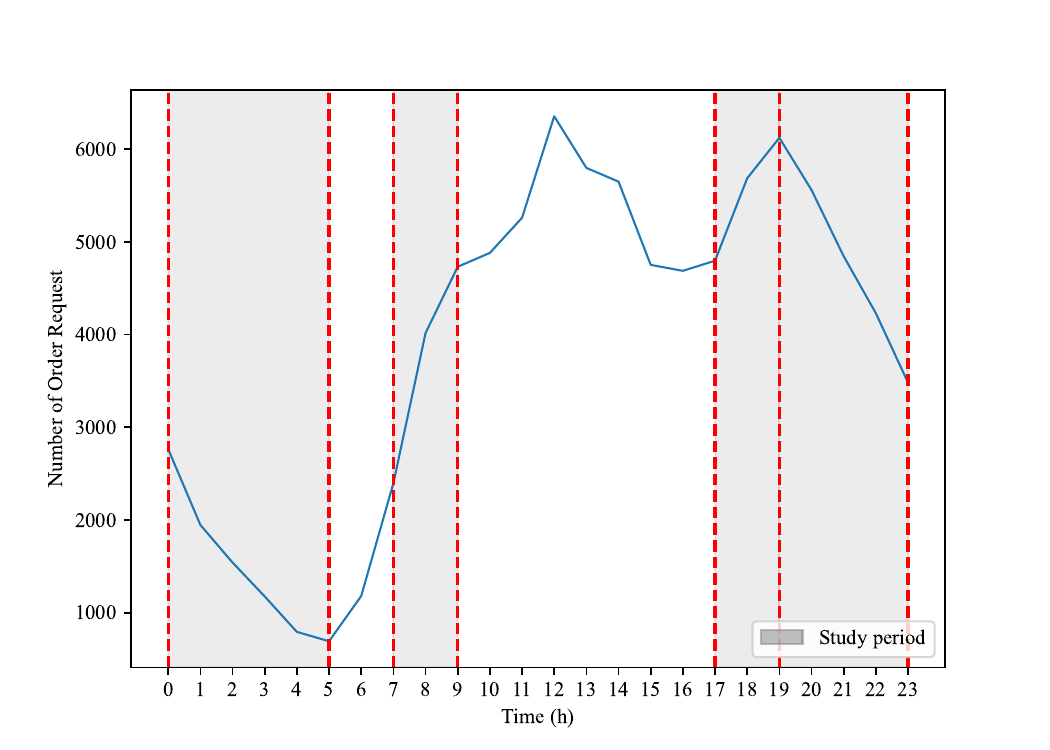}
\end{minipage}%
}%

\centering
\caption{The left and right figures respectively illustrate the hourly demand in Manhattan and Hong Kong, spanning from 0:00 to 23:00. The y-axis represents the total number of order requests appearing within each hour, while the x-axis denotes the time of day. The grey areas represent the study periods.}
\label{fig: temporal demand distribution}
\end{figure}

\begin{figure}[!t]
\centering
\subfigure[Demand at Manhattan]{
\begin{minipage}[t]{0.5\linewidth}
\centering
\includegraphics[trim=0cm 0cm 0cm 0cm,width=2.5in, height=2.5in]{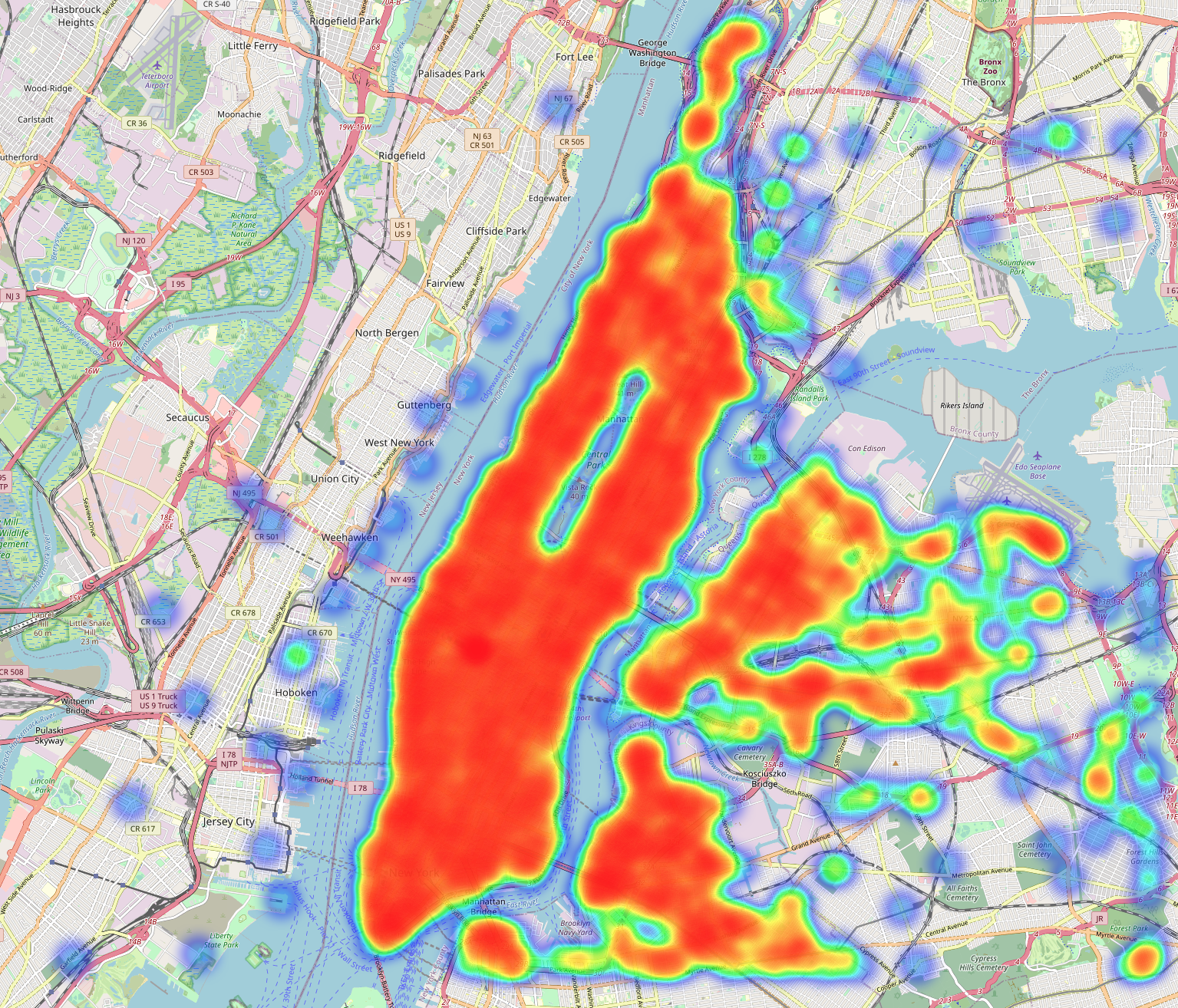}
\end{minipage}%
}%
\subfigure[Demand at Hong Kong Island]{
\begin{minipage}[t]{0.5\linewidth}
\centering
\includegraphics[trim=0cm 0cm 0cm 0cm,width=2.5in, height=2.5in]{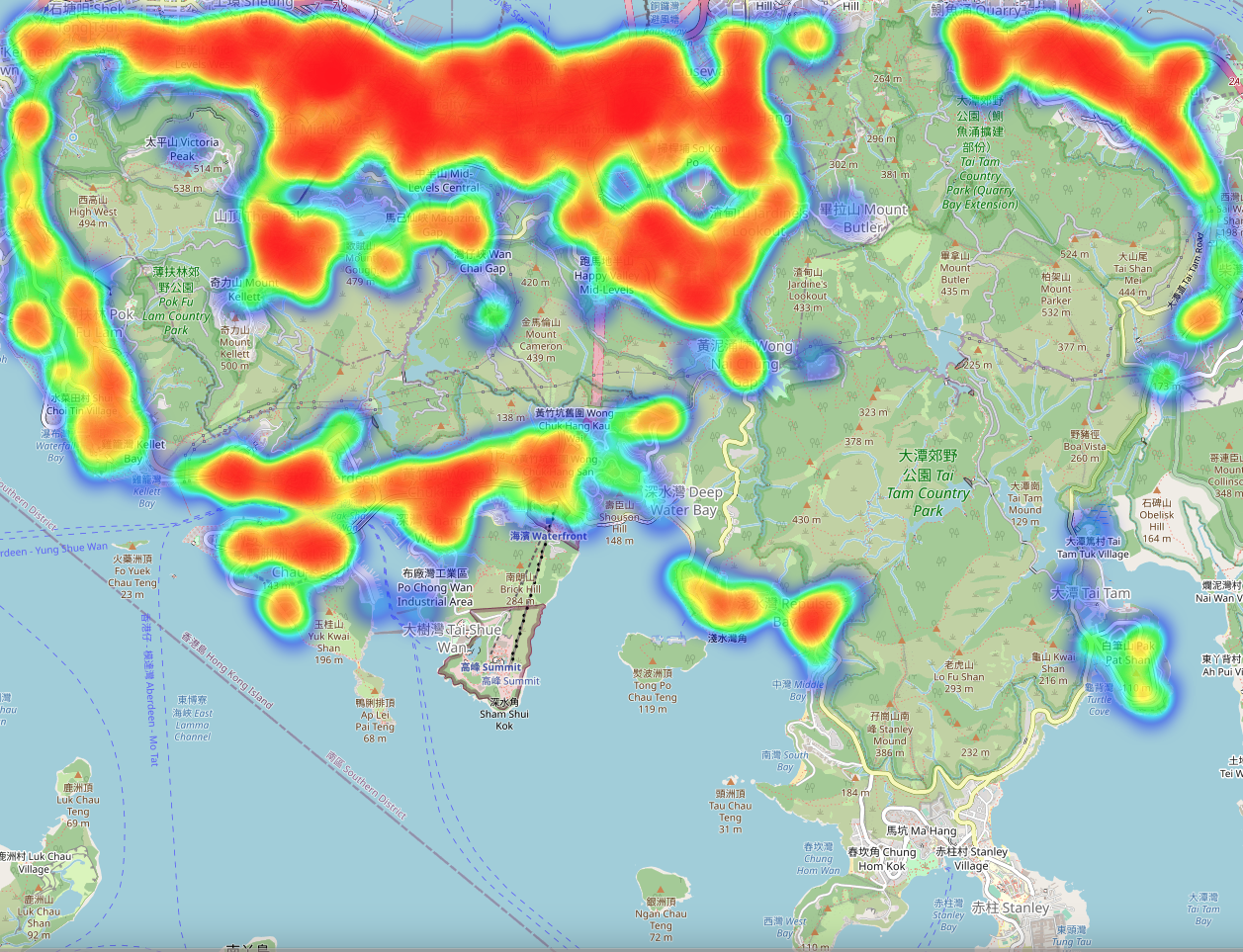}
\end{minipage}%
}%

\centering
\caption{The left and right figures respectively illustrate the spatial distribution of demand in Manhattan and Hong Kong.}
\label{fig: spatial demand distribution}
\end{figure}

In this study, we perform experiments using data from Manhattan and Hong Kong. We employ data from the Hong Kong Taxi Track Dataset. This dataset offers detailed GPS data, updated every two seconds, from 500 vehicles over two years. For the Manhattan experiment, we use data drawn from the NYC TLC Yellow Taxi dataset \footnote{https://www.nyc.gov/site/tlc/about/tlc-trip-record-data.page} for May 2015. The research regions in Manhattan and Hong Kong are depicted separately in Figure \ref{fig: research_area}. These regions are partitioned into 16 grids. Figure \ref{fig: temporal demand distribution} illustrates the demands of all the grids at various hours in both Manhattan and Hong Kong and shadow regions are our studied periods. It is evident from the figure that there are two peak periods in both regions, with the fewest orders occurring at 5 am. Figure \ref{fig: spatial demand distribution} provides a visual representation of the spatial distribution of demand in Manhattan and Hong Kong Island. The color scheme indicates order density, with darker red signifying higher density, and yellow representing lower density. To align the GPS data with the actual road network, we use a function provided by Osmnx \citep{boeing2017osmnx}, which matches real road nodes to specified longitude and latitude. The parameter settings for simulations in Manhattan and Hong Kong are detailed in Table \ref{tbl:settings for simulation}. Specifically, the simulation interval and the number of grids are identical for both regions, while other parameters differ. The hyperparameters used for TEB model training are outlined in Table \ref{tbl:hyperparameters of temporal model}.

\begin{table*}[!t]
    \centering
    \caption{The parameters setting of simulation in Manhattan and Hong Kong}
    \label{tbl:settings for simulation}
    \begin{tabular}{lll}
        \Xhline{0.5pt}
        Parameter & Hong Kong & Manhattan \\
        \Xcline{1-1}{0.5pt}
        \Xhline{0.5pt}
        Number of order & 44309 &  17558\\
        Number of driver & 200 &  500\\
        Number of grid  & 4*4 & 4*4\\
        Vehicle speed ($km/h$) & 20.6 & 22.79 \\
        Simulation interval ($s$) & 10 & 10\\
        \Xhline{0.5pt}
    \end{tabular} 
\end{table*}

\begin{table*}[!t]
    \centering
    \caption{Hyperparameters used for temporal model training}
    \label{tbl:hyperparameters of temporal model}
    \begin{tabular}{lll}
        \Xhline{0.5pt}
        Hyperparameter & &Value\\
        \Xcline{1-1}{0.5pt}
        \Xhline{0.5pt}
        Learning rate & &1e-3 \\
        Batch size & &1024 \\
        Decay factor & &0.1 \\
        Number of LSTM layers & &5\\
        Number of training steps to update the training strategy & &10\\
        Number of  time windows to record the historical losses & &10\\
        
        \Xhline{0.5pt}
    \end{tabular} 
\end{table*}

Table \ref{tbl: baseline models} itemizes all the baseline models with their corresponding descriptions. 'FR' signifies a uniform broadcasting radius across all grids for a given period, with its value varying from 1 to 4 in Hong Kong and 1 to 10 in Manhattan. For instance, a radius value of 1 is conveyed as 'FR-1'. To emulate the broadcasting process within the ride-sourcing market, we adapted a Python-coded platform \citep{feng2023multi}. The training and simulation phases were executed on a server furnished with an Intel(R) Core(TM) i9-10900X CPU, 128 GB RAM, and two Nvidia 4090 GPUs.
\begin{table}[!t]
    \centering
    \caption{Baseline models}
    \label{tbl: baseline models}
    \begin{tabular}{p{0.2\textwidth} p{0.7\textwidth}}
        \Xhline{0.5pt}
        Model & Explanation \\
        \Xcline{1-1}{0.5pt}
        \Xhline{0.5pt}
        FR-n & Fix radius as n, the broadcasting radius is fixed over the whole period over all the grids. \\
        LSTM  & LSTM model, \cite{hochreiter1997long} introduce the gate function, which can solve the problem with long-term time dependencies well. \\
        
        \Xhline{0.5pt}
    \end{tabular} 
\end{table}

\subsection{Evaluation Metrics and Results Analysis}
\begin{table*}[!b]
    \centering
    \caption{Simulation result in Manhattan}
    \label{tbl:simulation result manhattan}
    \begin{tabular}{llllll}
        \Xhline{0.5pt}
        Model & Training Strategy & OFR & DUR & PR (USD) & APD (km)\\
        \Xcline{1-6}{0.5pt}
FR-1 & - & 0.296 & 0.353 & 53991 & 1.231 \\		
FR-2 & - & 0.327 & 0.377 & 61194 & 2.152 \\	
FR-3 & - & 0.404 & 0.449 & 73286.5 & 3.150 \\	
FR-4 & - & 0.445 & 0.457 & 81250.5 & 4.223 \\
FR-5 & - & 0.597 & 0.522 & 108560.5 & 4.835 \\
FR-6 & - & 0.591 & 0.546 & 107012.5 & 5.426 \\
FR-7 & - & 0.579 & 0.567 & 105830.5 & 6.060 \\
FR-8 & - & 0.570 & 0.577 & 104073.5 & 6.579 \\
FR-9 & - & 0.566 & 0.592 & 102297 & 6.983 \\
FR-10 & - & 0.554 & 0.588 & 99713.5 & 7.417 \\
LSTM & WESM & 0.555 & 0.589 & 101779 & 7.138\\
TEB & WESM  & \textbf{0.675} & 0.44 & \textbf{116759.5} & 2.85\\
        \Xhline{0.5pt}
    \end{tabular} 
\end{table*}
\begin{table*}[!t]
    \centering
    \caption{Simulation result in Hong Kong}
    \label{tbl:simulation result hk}
    \begin{tabular}{llllll}
        \Xhline{0.5pt}
        Model & Training Strategy & OFR & DUR & PR (HKD) & APD (km)\\
        \Xcline{1-6}{0.5pt}
FR-1 & - & 0.333 & 0.223 & 674809.3 & 1.831 \\	
FR-2 & - & 0.353 & 0.224 & 717012.5 & 2.574 \\
FR-3 & - & 0.357 & 0.217 & 721971.6 & 3.323 \\
FR-4 & - & 0.358 & 0.214 & 723027.3  & 4.075 \\
LSTM & WESM  & 0.346 & 0.213 & 704423.7 & 2.786\\
TEB & WESM & 0.355 & 0.218 & \textbf{723034.6} & 2.803\\
        \Xhline{0.5pt}
    \end{tabular} 
\end{table*}

In our experiments, we evaluated metrics such as the Order Fulfillment Rate (OFR), Driver Utilization Rate (DUR), Platform Revenue (PR), and Average Pickup Distance (APD). These metrics serve as our optimizing objectives.

The simulation results from Hong Kong are displayed in Table \ref{tbl:simulation result hk}, which includes both fixed radii from 1 to 4 and dynamically adjusted radii using our temporal model. Notably, the Transformer-Encoder-Based (TEB) model for the e-hailing broadcasting system outperforms all other models, achieving a platform revenue (PR) that is 7.15\% higher than the model using 1 as the fixed radius. Although the PR of the FR-4 model is nearly equivalent to the TEB model, the latter significantly reduces the average pickup distance by 31.2\%, indicating a substantial advantage in terms of efficiency. While the DURs of FR-1 and FR-2 surpass the TEB model, their performance in other metrics falls short.

Transitioning to the Manhattan results (Table \ref{tbl:simulation result manhattan}), we examined fixed radii from 1 to 10 and dynamically adjusted radii using our temporal model. The TEB model stands out by achieving the highest OFR (0.675) and PR (116759.5). It also records a relatively higher DUR (0.44) and a lower APD (2.85), demonstrating superior performance with a 13\% improvement in OFR, 7.55\% in PR, and a reduction of 41\% in APD compared to the FR-5 model.

For baseline models utilizing a fixed radius, as the radius value increases, the APD also increases. However, the OFR, DUR, and PR initially increase but later fluctuate or even decrease. This suggests that neither the smallest nor the largest radius is the optimal matching radius.

The LSTM model, despite its satisfactory performance, does not match the efficiency and effectiveness of the TEB model in both Manhattan and Hong Kong. This indicates the superiority of the TEB model in managing the dynamics and extracting temporal patterns of ride-hailing services under the broadcasting mode in various urban environments.

The TEB model's superiority can be attributed to its ability to dynamically adjust the radius according to the spatial-temporal distribution of demand and supply. This ensures that drivers will not accept orders with excessively long pickup times, resulting in shorter distances to pick up orders. It also allows drivers to select higher-value orders, resulting in higher overall income.

These results underline the importance of using dynamic matching radii for ride-hailing systems. The TEB model, with its ability to adapt to different supply and demand conditions, hasshown significant potential for enhancing the efficiency of ride-hailing operations. Considering the real-world implications, the fluctuation of OFR, DUR, and PR might lead to inconsistent service experiences for both drivers and passengers. For instance, a decrease in OFR could mean that fewer orders are fulfilled, potentially leading to longer waiting times for passengers and lower revenue for drivers. Moreover, an increase in APD could result in longer pick-up times, negatively affecting service efficiency. Therefore, the ability of the TEB model to maintain consistent performance across these metrics could significantly improve the user experience and operational efficiency in ride-hailing services.

The adaptability of the TEB model to different supply and demand conditions offers a promising outlook for ride-hailing services. This flexibility could lead to more efficient resource allocation, resulting in improved service quality and profitability. For instance, during peak hours, the dynamic adjustment of the matching radius could ensure that orders are distributed more evenly among drivers, reducing order fulfillment times and improving driver utilization rates. In contrast, during off-peak hours, it could help to concentrate orders among fewer drivers, minimizing idle times and ensuring stable income for active drivers.

In conclusion, our experimental results provide strong evidence for the benefits of employing dynamic matching radii, as exemplified by the TEB model, in ride-hailing systems. This research offers valuable insights for both academics and practitioners in the field of ride-hailing operations and beyond.
{\renewcommand{\arraystretch}{1.2}
\begin{table}[!t]
    \centering
    \caption{Training strategies for ablation study}
    \label{tbl: baseline strategies}
    \begin{tabular}{p{0.2\textwidth} p{0.7\textwidth}}
        \Xhline{0.5pt}
        Strategy & Explanation \\
        \Xcline{1-1}{0.5pt}
        \Xhline{0.5pt}
FW & FW refers to Fixed Weight, a strategy that maintains a constant weight for all tasks throughout the training procedure.\\		
AM &  AM, standing for Adaptive Multi-task training strategy, operates in such a way that, at each training iteration, the cumulative historical loss for every task is computed. Subsequently, the parameters associated with the task exhibiting the maximum cumulative loss are updated.\\	 
WAM & WAM, an acronym for Weighted Adaptive Multi-task training strategy, is an extension of the AM approach. WAM employs a weight, determined by dividing the cumulative historical loss of each task by the total cumulative historical losses across all tasks, to concurrently update the parameters of all tasks.\\	
ESM &  ESM, short for Exponential Smoothing multi-task training strategy, is an extension of the AM strategy. However, it diverges from AM in its approach to calculating cumulative historical loss. Specifically, ESM takes into account that the impact of historical losses exponentially decays as the distance from the respective historical loss to the current training step increases.\\         
        \Xhline{0.5pt}
    \end{tabular} 
\end{table}
}
\subsection{Ablation Study}

In this section, we conduct an ablation study to explore the effects of the exponential smoothing decay and weighting mechanism in our proposed Weighted Exponential Smoothing Multi-task (WESM) learning strategy. For comparison, we introduce the Fixed Weight (FW), Adaptive Multi-task (AM), Weighted Adaptive Multi-task (WAM), and Exponential Smoothing (ESM) strategies. The details of these strategies and a comparison between them are outlined in Table \ref{tbl: baseline strategies}.

\begin{figure*}[!t]
  \begin{minipage}{1\textwidth}
    \centering
    \includegraphics[trim=5cm 0 4cm 0cm, width=1\textwidth,]{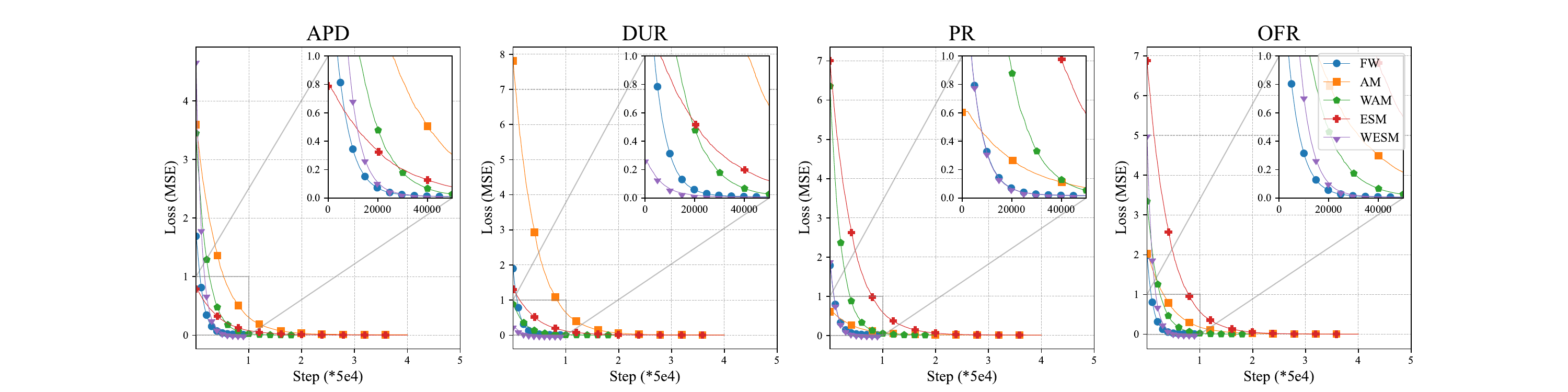}
    \caption{Test loss of each prediction task in Hong Kong}
    \label{fig:test_loss_hk}
  \end{minipage}
\end{figure*}

As shown in Figure \ref{fig:test_loss_hk} for the Hong Kong training results, the model's loss across all strategies converges to a notably low point. However, our WESM strategy demonstrates nearly faster convergence than the FW strategy for tasks related to predicting the driver utilization rate (DUR), platform revenue (PR), and order fulfillment rate (OFR). For example, after 5000 steps of training, the loss of using WESM to predict PR is only about half of the loss of using FW to predict PR. This faster convergence of the WESM strategy indicates its enhanced ability to quickly adapt and learn from the training data, potentially leading to improved model efficiency. Moreover, this may facilitate the prediction of these key performance indicators in a timely manner, thereby allowing for rapid adjustments in ride-hailing operations.

\begin{figure*}[!t]
  \begin{minipage}{1\textwidth}
    \centering
    \includegraphics[trim=5cm 0 4cm 0cm, width=1\textwidth,]{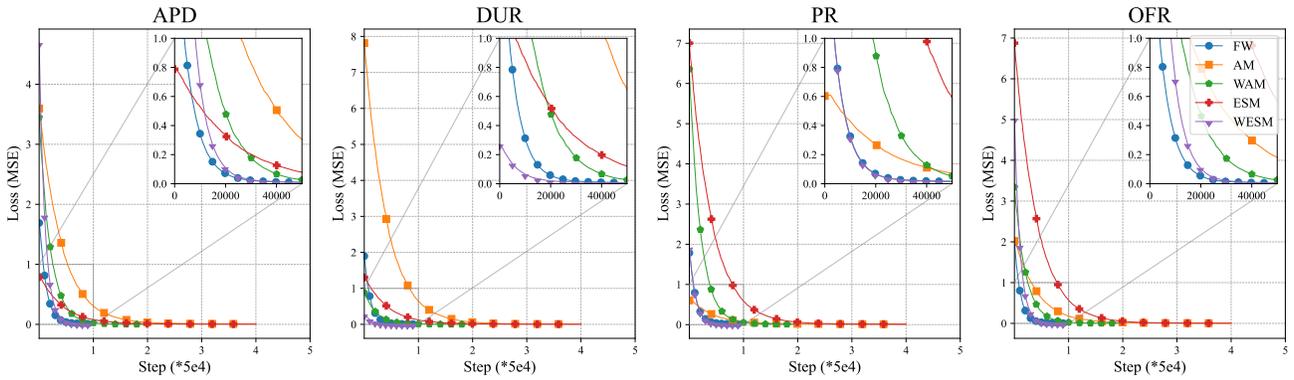}
    \caption{Test loss of each prediction task in Manhattan}
    \label{fig:test_loss_manhattan}
  \end{minipage}
\end{figure*}

\begin{table*}[!t]
    \centering
    \caption{Ablation studies in Hong Kong and Manhattan}
    \label{tbl:ablation study}
    \begin{tabular}{ll|llll|llll}
        \Xhline{0.5pt}
        \multirow{2}{*}{Model} & \multirow{2}{*}{Strategy} & \multicolumn{4}{c}{Hong Kong} & \multicolumn{4}{c}{Manhattan} \\
        \cmidrule{3-5} \cmidrule{6-10} 
         & & OFR & DUR & PR (HKD) & APD (km) & OFR & DUR & PR (USD) & APD (km) \\
        \Xcline{1-1}{0.5pt}
        \Xhline{0.5pt}		
TEB&FW  & 0.322 & 0.203 & 641486.3 & 2.24 & 0.562 & 0.621 & 100787 & 7.115\\
TEB&AM  & 0.348 & 0.216 & 708121.9 & 2.598 & 0.582 & 0.56 & 106985 & 5.849\\
TEB&WAM & 0.35 & 0.217 & 711215.5 & 2.71 & 0.554 & 0.594 & 99993 & 7.384\\
TEB&ESM  & 0.355 & 0.217 & 721896.5 & 2.79 & 0.569 & 0.587 & 98485 & 5.944 \\
TEB&WESM  &\textbf{0.355} & \textbf{0.218} & \textbf{723034.6} & 2.803  & \textbf{0.675} & 0.44 & \textbf{116759.5} & \textbf{2.85}\\
        \Xhline{0.5pt}
    \end{tabular} 
\end{table*}
Transitioning to the Manhattan experiment (Figure \ref{fig:test_loss_manhattan}), we observe that the training using the WESM strategy achieved faster convergence than the AM, WAM, and ESM strategies. This demonstrates the robustness of the WESM strategy across diverse urban environments. Despite the convergence speed of the WESM strategy for predicting average pick-up distance (APD), PR, and MR closely mirroring that of the FW strategy, the WESM strategy demonstrates quicker convergence in tasks predicting DUR. This suggests that the WESM strategy may be particularly beneficial for tasks with more complex or unpredictable patterns, such as DUR.

We validated the models trained using these strategies in our simulator for both Hong Kong and Manhattan. The results, displayed in Table \ref{tbl:ablation study}, show that the Transformer-Encoder-Based (TEB) model for the e-hailing broadcasting system trained with WESM outperforms other strategies in most scenarios. For instance, in Hong Kong, it attains the highest OFR, DUR, and PR. In Manhattan, it achieves the highest OFR, PR, and APD. The OFR value for the TEB model trained with the WESM strategy shows an improvement of 10.25\% and 20.11\% in Hong Kong and Manhattan, respectively, when compared to the TEB model trained with the FW strategy. These significant improvements underline the effectiveness of the WESM strategy in enhancing the performance of the TEB model in various metrics, thus highlighting its potential for improving ride-hailing operations across different urban environments.

The superior performance of the WESM strategy can be attributed to its adaptive nature and the exponential smoothing decay mechanism. The adaptive nature of WESM allows it to dynamically adjust the weights of different tasks based on their historical losses, enabling it to focus on tasks that require more learning. The exponential smoothing decay mechanism ensures that more recent losses have a higher impact on the weights, making the strategy more responsive to recent changes in the data. These characteristics make the WESM strategy more efficient and effective for multi-task learning in the context of ride-hailing services.

In conclusion, our ablation study provides strong evidence for the benefits of employing the WESM strategy in multi-task learning for ride-hailing services. This strategy not only fosters quicker convergence in multiple task predictions but also improves the overall performance of the model across varied urban environments.

Further research could focus on refining the WES strategy or developing new strategies that build upon its strengths. For example, incorporating other types of decay mechanisms or using a different approach to weight adaptation could potentially enhance its performance. It would also be beneficial to test the WESM strategy in other types of multi-task learning applications, beyond ride-hailing, to assess its generalizability and adaptability. Moreover, future work could delve deeper into understanding the characteristics of the tasks where the WESM strategy performs particularly well. This analysis could provide insights into the types of tasks that benefit most from this strategy, which could guide the design of new strategies and help optimize multi-task learning systems.

Lastly, while our ablation study has focused on comparing the performance of different strategies in a controlled environment, it would be valuable to validate these findings in a real-world setting. Implementing these strategies in a live ride-hailing system and measuring their impact on key operational metrics would provide further evidence for the effectiveness of the WESM strategy and help gauge its practical implications.

\section{Conclusion}
This paper investigates the dynamic adjustment of matching radii in the context of the broadcasting mode within the ride-hailing market. We introduce the Dynamic Broadcasting Radii Adjustment System (DBRAS) to enhance overall performance metrics while respecting driver autonomy in selecting preferred orders. The system incorporates a Transformer-Encoder-Based (TEB) model to predict the values of four performance metrics (Order Fulfillment Rate (OFR), Platform Revenue (PR), Driver Utilization Rate (DUR), and Average Pickup Distance (APD)) across different times and regions. Subsequently, near-optimal broadcasting radii are determined through an algorithm that integrates these predicted metric values. To address the challenge of predicting multiple performance metrics, we propose the Weighted Exponential Smoothing Multi-task (WESM) learning strategy to improve prediction accuracy. The effectiveness of our proposed temporal model and multi-task learning strategies is validated through comprehensive experiments. Our findings indicate that when demand is high, the optimal matching radius is not large; it merely needs to ensure that each driver can evenly be distributed with some available orders for selection. Conversely, when demand is considerably low, the approximate optimal radius is relatively large, but only needs to ensure that drivers have orders to choose from within this radius. This approach avoids excessive pickup distances without requiring an excessively large radius value.

In future research, there is considerable scope for further optimization of the broadcasting mode within the ride-hailing market by developing innovative algorithms. This may involve devising novel broadcasting strategies, such as adjusting the broadcasting frequency or tailoring broadcasting orders based on individual driver profiles, to provide a more personalized and efficient service delivery. Investigating mixed markets that combine the broadcasting and the dispatching modes also presents a promising research direction, for example, identifying the optimal ratio between broadcasting and dispatching to maximize overall market performance. Another area of potential focus is the integration of radius optimization with the passenger-driver matching process. This joint optimization could potentially augment the efficiency and effectiveness of ride-hailing services, thereby establishing a new benchmark for the industry.

\section{Acknowledgements}
This research is supported by the Smart Traffic Fund(PSRI/29/2201/PR) of the Hong Kong SAR Government. We are also very grateful to eTaxi (\href{https://etaxi.com.hk/}{https://etaxi.com.hk/}) for providing us with sample data.

\newpage

\appendix
\section*{Appendix}
\noindent\textbf{Multi-Layer Perceptron:} MLP consists of multiple layers of nodes in a directed graph, where each layer is fully connected to the next one. The representation of an MLP with one hidden layer is the following:
\begin{align*}
\mathbf{H} &= f(\mathbf{X} \mathbf{W}_1 + \mathbf{b}_1), \\
\mathbf{Y} &= \mathbf{H} \mathbf{W}_2 + \mathbf{b}_2,
\end{align*}
In the above equations:
\begin{itemize}
    \item $\mathbf{X}$ is the input.
    \item $\mathbf{W}_1$ and $\mathbf{W}_2$ are the weights for the first and second layers, respectively.
    \item $\mathbf{b}_1$ and $\mathbf{b}_2$ are the bias terms for the first and second layers, respectively.
    \item $f$ is the activation function.
    \item $\mathbf{H}$ is the output of the hidden layer.
    \item $\mathbf{Y}$ is the final output.
\end{itemize}
\textbf{Layer Normalization:} Layer normalization is one type of normalization technique, that performs normalization for each sample individually, used widely in natural language processing.
\begin{align*}
\mu &= \frac{1}{H}\sum_{i=1}^{H} x_i, \\
\sigma &= \sqrt{\frac{1}{H}\sum_{i=1}^{H}(x_i - \mu)^2 + \epsilon}, \\
\text{LayerNorm}(\mathbf{X}) &= \gamma \frac{\mathbf{X} - \mu}{\sigma} + \beta,
\end{align*}
In the above equations:
\begin{itemize}
    \item $\mathbf{X}$ is the input.
    \item $\mu$ is the mean of the elements of $\mathbf{X}$.
    \item $\sigma$ is the standard deviation of the elements of $\mathbf{X}$.
    \item $\gamma$ and $\beta$ are learnable parameters.
    \item $\epsilon$ is a small constant for numerical stability.
    \item $H$ is the size of the hidden layer.
    \item $\text{LayerNorm}(\mathbf{X})$ is the output after applying layer 
\end{itemize}
\textbf{Self-Attention:} Self-attention mechanism calculates the relevance of each input concerning the other inputs. It computes the weights (attention scores) using the query ($Q$), key ($K$), and value ($V$) vectors for each input. The output is a weighted sum of the value vectors.
\begin{align*}
\mathbf{Q} &= \mathbf{X} \mathbf{W}_{Q}, \\
\mathbf{K} &= \mathbf{X} \mathbf{W}_{K}, \\
\mathbf{V} &= \mathbf{X} \mathbf{W}_{V}, \\
\mathbf{Z} &= \text{softmax}\left(\frac{\mathbf{Q} \mathbf{K}^T}{\sqrt{d_k}}\right) \mathbf{V},
\end{align*}
In the above equations:
\begin{itemize}
    \item $\mathbf{X}$ is the input.
    \item $\mathbf{W}_{Q}$, $\mathbf{W}_{K}$, and $\mathbf{W}_{V}$ are the weight matrices for the query, key, and value, respectively.
    \item $\mathbf{Q}$, $\mathbf{K}$, and $\mathbf{V}$ are the query, key, and value vectors, respectively.
    \item $\mathbf{Z}$ is the output after applying the softmax function to the scores and multiplying by the value vector.
    \item $d_k$ is the dimension of the key vector.
\end{itemize}
\newpage

\bibliography{reference}
\biboptions{authoryear} 
\end{document}